\documentclass[lettersize,journal]{IEEEtran}

\usepackage[table]{xcolor}   % 支持 \rowcolor, \cellcolor
\usepackage{tabularx}        % 支持自动调整列宽的 tabularx 表格环境
\usepackage{booktabs}        % 用于 \toprule, \midrule 等
\usepackage{multirow}        % 多行单元格
\usepackage{amsmath}         % 数学符号和公式
\usepackage{makecell}

\usepackage{amsfonts}
\usepackage{algorithmic}
\usepackage{algorithm}
\usepackage{array}
\usepackage{textcomp}
\usepackage{stfloats}
\usepackage{url}
\usepackage{verbatim}
\usepackage{graphicx}
\usepackage{cite}
\usepackage[colorlinks,
            linkcolor=blue,
            citecolor=blue]{hyperref}
            
\usepackage[hyphenbreaks]{breakurl}
\usepackage{siunitx} % for better number alignment
\ifCLASSOPTIONcompsoc
  \usepackage{subfig}
\else
  \usepackage[caption=false, font=footnotesize]{subfig}
\fi

\def \NOTE [#1]{\textcolor{blue}{(\textit{#1})}}
\usepackage{xspace}
\usepackage{color}
\usepackage{multirow}
\usepackage{ifthen}
\usepackage{makecell}

\long\def\ignorethis#1{}

\definecolor{gray}{rgb}{0.35,0.35,0.35}
\definecolor{MyBlue}{rgb}{0,0.2,0.8}
\definecolor{MyRed}{rgb}{0.8,0.2,0}
\definecolor{MyGreen}{rgb}{0.0,0.5,0.1}
\definecolor{MyGray}{rgb}{0.4,0.4,0.4}

\newlength\paramargin
\newlength\figmargin
\newlength\subfigmargin
\newlength\secmargin
\newlength\subsecmargin
\newlength\tabmargin
\newlength\eqmargin

\usepackage{array}
\newcolumntype{L}[1]{>{\raggedright\let\newline\\\arraybackslash\hspace{0pt}}m{#1}}
\newcolumntype{C}[1]{>{\centering\let\newline\\\arraybackslash\hspace{0pt}}m{#1}}
\newcolumntype{R}[1]{>{\raggedleft\let\newline\\\arraybackslash\hspace{0pt}}m{#1}}

\def\ie{i.e.,~}
\def\eg{e.g.,~}

\newcommand{\secref}[1]{Section~\ref{sec:#1}}

\newcommand{\figref}[1]{Fig.~\ref{fig:#1}}
\newcommand{\tabref}[1]{Table~\ref{tab:#1}}

\newcommand{\Paragraph}[1]{\noindent\textbf{#1}}

\usepackage{pifont}
\usepackage{bbding}
\usepackage{float}
\usepackage{cite}
\usepackage{paralist}
\definecolor{mycolor_blue}{RGB}{231,239,250}
\definecolor{mycolor_green}{RGB}{230,247,224}
\definecolor{mycolor_gray}{RGB}{236,236,236}
\definecolor{pearDark!20}{RGB}{212,230,241}

\newcolumntype{P}[1]{>{\raggedright\arraybackslash}p{#1\linewidth}} % 自动适应文本宽度
\begin{document}

%\title{\title{Dual Latent Experts: Decoder-side Fusion for the Fidelity–Perception Tradeoff in Learned Image Compression}
%}PlugMoE: 
%\title{Dual Latent Mixture of Experts at Decoder-Side for Fidelity-Perception Balance in Learned Image Compression}
%\title{Decoder-Side Dual-Latent Mixture-of-Experts for Distortion–Perception Balancing in Learned Image Compression}
%\title{Balancing Fidelity and Perception via Dual-Latent Mixture of Decoder Experts in Learned Image Compression}
%\title{Dual-Latent Mixture of Decoder Experts for Fidelity–Perception Balanced Image Compression}

\title{Dual-Latent Collaborative Decoding for Fidelity–Perception Balanced Image Compression}

\author{Qi Mao, \textit{Member, IEEE}, Zijian Wang, Zhengxue Cheng, \textit{Member, IEEE}, Lingyu Zhu, Siwei Ma,  \textit{Fellow, IEEE}
\thanks{Qi Mao and Zijian Wang are with the School of Information and Communication Engineering and the State Key Laboratory of Media Convergence and Communication, Communication University of China, Beijing 100024, China (E-mail:  {qimao, wangzijian}@cuc.edu.cn) 
(Corresponding author: Qi Mao and Zhengxue Cheng).
\\
Zhengxue Cheng is with the School of Information Science and Electronic Engineering, Shanghai Jiao Tong University (E-mail: zxcheng@sjtu.edu.cn)
.
\\
Lingyu Zhu is with the Department of Computer Science, City University of Hong Kong (E-mail:lingyzhu-c@my.cityu.edu.hk).
\\
Siwei Ma is with the State Key Laboratory of Multimedia information
Processing, School of Computer Science, Peking University, Beijing 100871, China (E-mail:swma@pku.edu.cn).}}

\markboth{Journal of \LaTeX\ Class Files,~Vol.~14, No.~8, August~2021}%
{Shell \MakeLowercase{\textit{et al.}}: A Sample Article Using IEEEtran.cls for IEEE Journals}

\maketitle
%\IEEEpeerreviewmaketitle

\begin{abstract}
Learned image compression (LIC) increasingly requires reconstructions that balance distortion fidelity and perceptual realism across a wide range of bitrates.
However, most existing methods still rely on a single compressed latent representation to simultaneously carry structural details, semantic cues, and perceptual priors, requiring the same latent representation to serve multiple, potentially conflicting roles. 
This tension becomes evident across different latent paradigms: scalar-quantized (SQ) continuous latents provide rate-scalable fidelity but tend to lose perceptual details at low rates, while vector-quantized (VQ) discrete tokens preserve compact semantics cues but suffer from limited structural fidelity and bitrate scalability.
To address this issue, we propose \emph{Mixture of Decoder Experts} (MoDE), a dual-latent collaborative decoding framework that decomposes reconstruction responsibilities across complementary latent paradigms. 
Specifically, MoDE treats the SQ branch as a fidelity-oriented expert and the VQ branch as a perception-oriented expert, and coordinates them through two decoder-side modules: \emph{Expert-Specific Enhancement} (ESE), which preserves branch-specific expert references, and \emph{Cross-Expert Modulation} (CEM), which enables selective complementary transfer during reconstruction. 
The resulting framework supports selective cross-latent collaboration under a shared dual-stream bitstream and enables both fidelity-anchored and perception-anchored decoding.
Extensive experiments demonstrate that MoDE achieves a more \emph{favorable fidelity–perception balance} than representative distortion-oriented, perception-oriented, generative, and dual-latent baselines across a wide bitrate range, highlighting decoder-side expert collaboration as an effective design for wide-range fidelity–perception balanced LIC.
\end{abstract}

\begin{IEEEkeywords}
learned image compression, fidelity--perception trade-off, mixture of decoder experts, perceptual compression
\end{IEEEkeywords}

\section{Introduction}

\IEEEPARstart{L}{earned} image compression (LIC) has achieved remarkable progress in rate--distortion optimization, yet modern image compression increasingly requires reconstructions that balance distortion fidelity and perceptual realism across a wide range of bitrates.
A common design choice is to encode an image into a single compressed latent representation and require it to simultaneously support entropy-efficient coding, structural recovery, semantic preservation, and perceptually plausible reconstruction.
However, these objectives impose multiple, potentially conflicting roles on the same latent space.
As a result, a single-latent design is often forced to favor one reconstruction objective at the expense of another, leading to a persistent fidelity--perception tension in learned compression.

Most distortion-oriented LIC methods~\cite{balle2017end, balle2018variational, minnen2018joint, cheng2019energy, ma2019iwave, cheng2020learned, fu2023learned, he2022elic, akbari2021learned} adopt scalar-quantized (SQ) continuous latents, which are well suited to fine-grained rate control, and scalable structural fidelity. 
However, under tight rate budgets, SQ-based latents tend to suppress subtle textures and semantic details, often producing overly smooth reconstructions. 
In contrast, perception-oriented compression methods~\cite{el2022image, mao2024extreme, xue2024unifying, jia2024generative} increasingly exploit vector-quantized (VQ) discrete tokens to leverage compact semantic cues and strong generative priors~\cite{van2017neural, esser2021taming, chang2022maskgit}. 
Such representations are particularly attractive at very low bitrates, but they do not naturally guarantee accurate recovery of fine structures and often offer less flexible bitrate scalability. 
Taken together, these observations suggest that SQ and VQ latent paradigms are better suited to complementary reconstruction roles, rather than a single overloaded latent representation.

\begin{figure*}[!t]
    \includegraphics[width=1.0\linewidth]{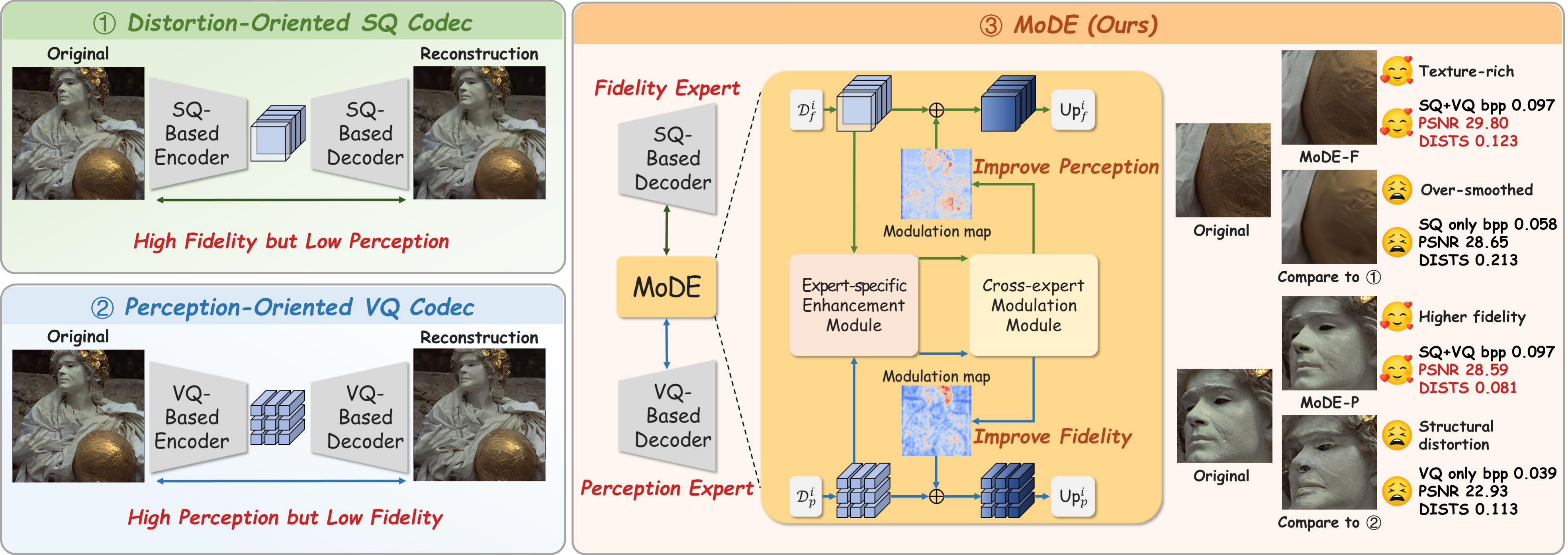}
\vspace{- 2mm}
\caption{
\textbf{MoDE: Dual-latent collaborative decoding for fidelity--perception balanced image compression.}
(1) Distortion-oriented SQ codecs preserve structural fidelity but tend to over-smooth at low-rate reconstructions.
(2) Perception-oriented VQ codecs preserve compact semantics at ultra-low rates but have weaker structural fidelity and bitrate scalability.
(3) MoDE coordinates frozen SQ and VQ decoders as fidelity and perception experts through \emph{ESE} and \emph{CEM} under a shared dual-stream bitstream, enabling both \textbf{MoDE-F} and \textbf{MoDE-P} with explicit fidelity--perception trade-offs.
Bitrates denote total bpp over the SQ and VQ streams.
}
\vspace{- 6mm}
\label{fig:teaser}
\end{figure*}

Building on this complementarity, several recent studies~\cite{lu2024hybridflow, iwai2024dual, xue2025dlf} have explored hybrid or multi-branch compression frameworks for combining heterogeneous latent representations.
Typically, such complementarity is realized through tightly coupled fusion, prediction, conditioning, or joint optimization, effectively merging different representations into a shared reconstruction pathway rather than assigning them distinct reconstruction responsibilities and preserving their native specialization.
Such tightly coupled interaction can weaken the native specialization of each branch, making stable fidelity--perception behavior harder to maintain across a wide bitrate range.
This leaves an important question open: \emph{how can heterogeneous compressed representations be coordinated such that one branch remains structurally faithful, the other contributes perceptual realism, and both remain effective across diverse bitrate regimes?}

To address this question, we adopt a decoder-side collaborative decoding perspective.
The key idea is to treat heterogeneous compressed representations not as features to be merged into a shared reconstruction pathway, but as complementary decoder experts with distinct reconstruction responsibilities.
As illustrated in \figref{teaser}, we propose \textbf{\emph{MoDE}}, a dual-latent collaborative decoding framework for fidelity--perception balanced image compression.
Specifically, MoDE treats the SQ branch as a fidelity-oriented expert and the VQ branch as a perception-oriented expert.
It then coordinates the two experts at the decoder side through an \textbf{\emph{Expert-Specific Enhancement (ESE)}} module and a \textbf{\emph{Cross-Expert Modulation (CEM)}} module, where ESE preserves branch-specific expert references and CEM enables selective complementary transfer between experts during reconstruction.
This design supports selective cross-latent collaboration under a shared dual-stream bitstream and enables both fidelity-anchored and perception-anchored decoding within a unified framework, yielding explicit fidelity--perception trade-offs across bitrate regimes.
Extensive experiments demonstrate that MoDE achieves a more \emph{favorable fidelity--perception balance} than representative distortion-oriented, perception-oriented, generative, and dual-latent baselines across a wide bitrate range.

In summary, our contributions are as follows:
\begin{compactitem}
\item We identify the single-latent overload issue in learned image compression and formulate it as a latent responsibility decomposition problem, thereby motivating explicit fidelity- and perception-oriented decoder roles.
\item We propose \textbf{MoDE}, a dual-latent collaborative decoding framework that coordinates SQ and VQ representations as complementary decoder experts, together with \textbf{ESE} for preserving expert specialization and \textbf{CEM} for enabling selective cross-expert collaboration.
\item We validate MoDE across multiple datasets, bitrate regimes, and representative codec paradigms, showing that it achieves a more favorable fidelity--perception balance than representative distortion-oriented, perception-oriented, generative, and dual-latent baselines over a wide bitrate range.
\end{compactitem}

\section{Related Work}
\label{sec:related_work}

\Paragraph{Learned Image Compression.}
LIC replaces hand-crafted coding modules with end-to-end optimized neural architectures and has achieved strong rate--distortion performance~\cite{balle2017end, balle2018variational, minnen2018joint, ma2019iwave, cheng2020learned, cheng2019energy, he2022elic, fu2023learned}.
Early works model continuous latents with scalar quantization (SQ) and entropy models for end-to-end RD optimization~\cite{balle2017end}, and subsequent advances improved entropy modeling via hyperpriors and context models~\cite{balle2018variational, minnen2018joint, ma2019iwave, cheng2019energy}.
Efficient context designs, such as checkerboard or grouped context modeling, reduce autoregressive complexity~\cite{he2022elic}, while mixture-based likelihood models further improve RD efficiency~\cite{cheng2020learned, fu2023learned}.
Despite these advances, distortion-oriented LIC mainly optimizes pixel-level fidelity and often produces over-smoothed reconstructions at low bitrates, motivating perceptual compression.

\begin{table*}[!t]
\centering
\caption{\textbf{Comparison of representative neural image codec paradigms.}
MoDE differs from prior dual-latent fusion methods by coordinating heterogeneous SQ/VQ decoders through decoder-side collaboration, rather than fusion or conditioning alone.}
\vspace{-2mm}
\label{tab:related_comparison}
\resizebox{\textwidth}{!}{%
    \begin{tabular}{l l l cc cc ccc c}
    \toprule
    \multirow{2}{*}{\textbf{Codec family}} 
    & \multirow{2}{*}{\textbf{Latent}} 
    & \multirow{2}{*}{\textbf{Repr. method}} 
    & \multicolumn{2}{c}{\textbf{Quantization space}} 
    & \multicolumn{2}{c}{\textbf{Main objective}} 
    & \multicolumn{3}{c}{\textbf{Support bitrate regime}}
    & \multirow{2}{*}{\textbf{Latent coordination}} \\
    \cmidrule(lr){4-5} \cmidrule(lr){6-7} \cmidrule(lr){8-10} 
    & & 
    & \textbf{Scalar} 
    & \textbf{Vector} 
    & \textbf{Fidelity} 
    & \textbf{Perception} 
    & \textbf{Ultra-low($<$0.05)} 
    & \textbf{Med.-low(0.05--0.2)} 
    & \textbf{Med./high(0.2--1)}
    & \\
    % & -- \\
    % & \textbf{Sel. modu.} \\
    % & & 
    % & 
    % & 
    % & 
    % & 
    % & \textbf{($<$0.05)} 
    % & \textbf{(0.05--0.2)} 
    % & \textbf{(0.2--1)} 
    % & \\
    \midrule
    Distortion-oriented SQ 
    & Single latent 
    & ELIC~\cite{he2022elic} 
    & \checkmark 
    & \ding{53} 
    & \checkmark 
    & \ding{53} 
    & \checkmark 
    & \checkmark 
    & \checkmark
    % & -- 
    & \ding{53}\\
    
    Perception-oriented SQ 
    & Single latent 
    & MS-ILLM~\cite{muckley2023improving} 
    & \checkmark 
    & \ding{53}
    & \ding{53}
    & \checkmark 
    & \checkmark
    & \checkmark 
    & \checkmark
    % & -- 
    & \ding{53} \\
    
    Perception-oriented VQ 
    & Single latent 
    & Finetuned-VQGAN~\cite{mao2024extreme} 
    & \ding{53} 
    & \checkmark 
    & \ding{53} 
    & \checkmark 
    & \checkmark 
    & \ding{53} 
    & \ding{53}  
    % & -- 
    & \ding{53} \\
    
    Dual-latent fusion 
    & Dual latent 
    & DC-VIC~\cite{iwai2024dual} / DLF~\cite{xue2025dlf} 
    & \checkmark 
    & \checkmark 
    & \ding{53} 
    & \checkmark 
    & \checkmark 
    & \checkmark 
    & \ding{53} 
    % & -- 
    & Fusion \\
    
    \textbf{MoDE} 
    & \textbf{Dual latent} 
    & \textbf{Ours} 
    & \textbf{\checkmark} 
    & \textbf{\checkmark} 
    & \textbf{\checkmark} 
    & \textbf{\checkmark} 
    & \textbf{\checkmark} 
    & \textbf{\checkmark} 
    & \textbf{\checkmark}
    % & \textbf{\checkmark}
    & \textbf{Decoder-side collaboration} \\
    \bottomrule
    \end{tabular}
}
\vspace{-5 mm}
\end{table*}

\Paragraph{Perceptual Learned Image Compression.}
Perceptual LIC aims to improve subjective visual quality, particularly at low bitrates where distortion-oriented optimization may lose texture and semantic details.
A common strategy is to augment RD training with perception-aligned objectives, such as feature-space perceptual losses~\cite{zhang2018unreasonable} and adversarial losses~\cite{goodfellow2014generative}.
GAN-based LIC was introduced by Agustsson \textit{et al.}~\cite{agustsson2019generative}, and subsequent methods further improve low-bitrate perceptual quality via conditional discriminators and training strategies~\cite{mentzer2020high, muckley2023improving}.
Beyond fixed loss weighting, Agustsson \textit{et al.}~\cite{agustsson2023multi} introduce learnable balancing to modulate the distortion--perception trade-off during training, while MS-ILLM~\cite{muckley2023improving} adopts a non-binary discriminator for enhanced perceptual quality.

Most of these perceptual codecs still rely on SQ-based continuous latents.
Although perceptual objectives can improve visual realism, they do not remove the reliance on entropy-constrained continuous latents, which may still provide limited semantic capacity under extremely tight rate budgets.
Inspired by VQ-based generative modeling, recent works~\cite{mao2024extreme, xue2024unifying, jia2024generative} explore tokenized perceptual compression based on VQGAN-style codebooks.
By transmitting discrete codebook indices, these methods represent high-level semantics compactly at very low bit costs.
However, compact tokenization can discard fine-grained fidelity information and degrade structural details.
These complementary strengths and limitations motivate dual-latent designs that combine tokenized perceptual coding with fidelity-oriented LIC components.

\Paragraph{Dual-Latent Learned Image Compression.}
Recent dual-latent codecs combine VQ-based tokenization with fidelity-oriented LIC, often through dual-pathway modeling or auxiliary conditioning. \tabref{related_comparison} summarizes the main design differences among representative codec paradigm.
HybridFlow~\cite{lu2024hybridflow} uses a pretrained LIC model to guide token prediction and reconstruction in the VQ stream.
DC-VIC~\cite{iwai2024dual} conditions a pretrained VQGAN decoder with auxiliary streams for token prediction and feature modulation to balance perceptual quality and fidelity.
DLF~\cite{xue2025dlf} decomposes latents into semantic and detail components and fuses them through interaction.

These methods demonstrate the value of heterogeneous latent representations, but they mainly exploit complementarity through prediction, conditioning, or feature fusion.
In contrast, our goal is not to simply merge SQ and VQ representations.
We treat the SQ and VQ decoders as specialized reconstruction experts with different responsibilities: a fidelity-oriented expert for scalable structural reconstruction and a perception-oriented expert for semantic realism under tight budgets.
MoDE then coordinates these experts at the decoder side while preserving their native specialization.

\Paragraph{Mixture-of-Experts for Vision Models.}
Mixture-of-Experts (MoE) was popularized as a conditional computation strategy, where only part of a model is activated for each input~\cite{shazeer2017outrageously}.
In vision, V-MoE~\cite{riquelme2021scaling} introduces sparse expert routing into Vision Transformers.
These works primarily use experts for capacity scaling or conditional routing.
In contrast, we introduce \emph{Mixture of Decoder Experts (MoDE)} for learned image compression, where pretrained SQ- and VQ-based decoders are treated as complementary experts with distinct fidelity- and perception-oriented reconstruction roles, and are coordinated through decoder-side collaborative decoding.

\section{Motivation}
\label{sec:motivation}

\subsection{Latent-Paradigm Evidence for Single-Latent Overload}
\label{sec:rdp}
\label{sec:latent_paradigms}

The rate--distortion--perception (RDP) framework~\cite{blau2019rethinking} formalizes the well-known tension among rate efficiency, distortion fidelity, and perceptual realism:
\begin{equation}
\begin{aligned}
R(D,P)=\min_{p(\hat{x}|x)} I(X;\hat{X})
\quad \text{s.t.} \quad
&\mathbb{E}[\Delta(X,\hat{X})]\le D,\\
&d(p_X,p_{\hat{X}})\le P,
\end{aligned}
\label{eq:rdp}
\end{equation}
where $d(\cdot,\cdot)$ denotes a distance between the source and reconstruction distributions.
However, in practice, adjusting the objective weights of a single codec mainly changes its optimization preference rather than removing the representational burden placed on the compressed latent, which is still expected to simultaneously carry rate-efficient information, distortion-faithful structure, and perception-relevant cues.

\begin{figure}[!t]
    \includegraphics[width=1.0\linewidth]{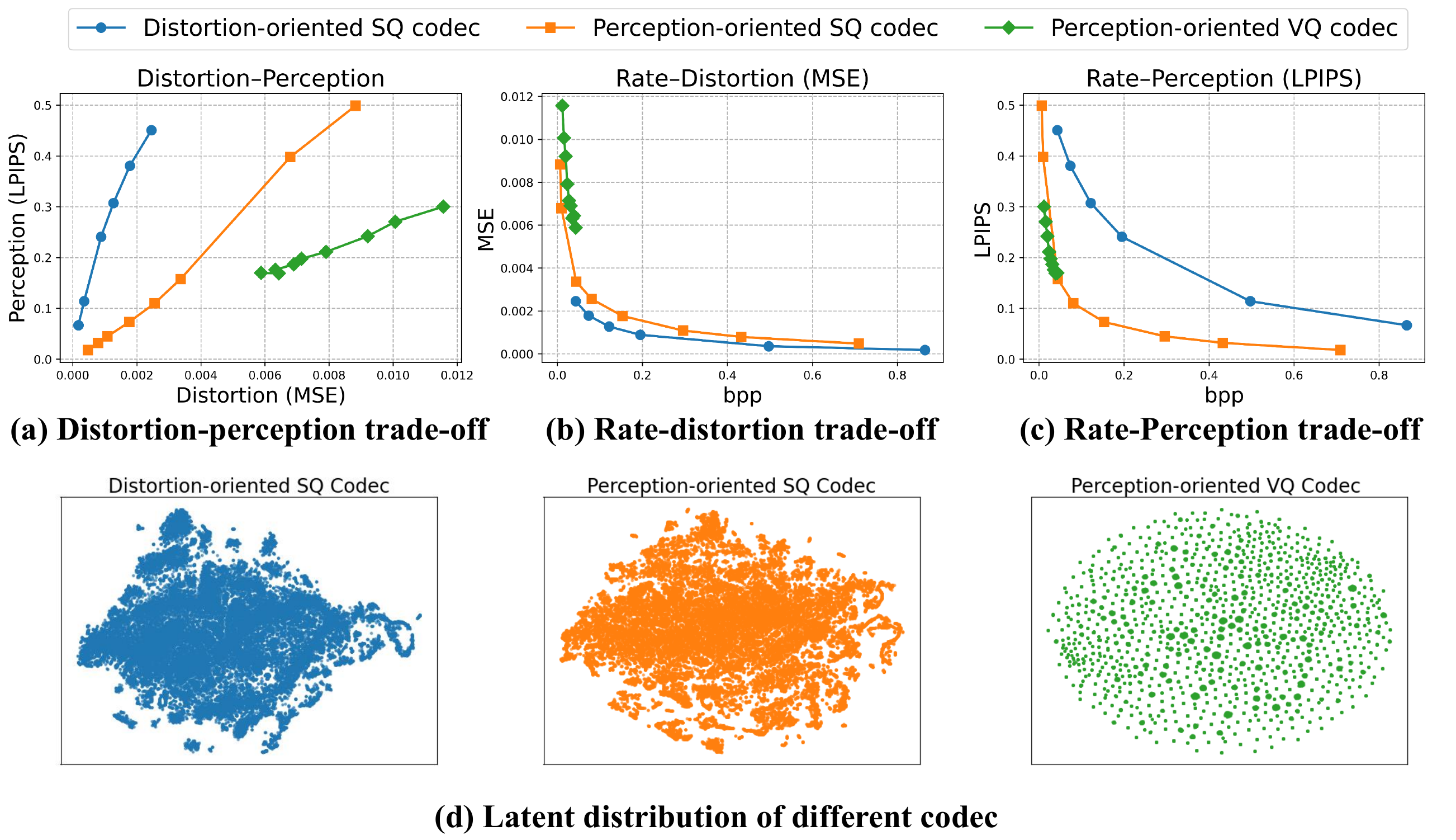}
\caption{\textbf{Latent-paradigm evidence under fidelity--perception tension.}
(a)--(c) SQ- and VQ-based codecs occupy different regions of the distortion--perception, rate--distortion, and rate--perception spaces.
(d) Their latent distributions exhibit different capacity allocation patterns, revealing distinct representational characteristics across latent paradigms.}
\label{fig:rdp}
\vspace{-5 mm}
\end{figure}

Fig.~\ref{fig:rdp} makes this single-latent tension concrete from a latent-paradigm perspective.
We compare three representative models: the distortion-oriented SQ codec ELIC~\cite{he2022elic}, the perceptual SQ codec MS-ILLM~\cite{muckley2023improving}, and the perception-oriented VQ codec Fine-tuned VQGAN~\cite{mao2024extreme}.
Here, \emph{SQ-based} denotes codecs with continuous latents and scalar quantization, while \emph{VQ-based} denotes tokenized perceptual codecs with discrete codebook indices and stronger generative priors. This distinction concerns the overall latent paradigm rather than the quantizer alone.

As shown in Fig.~\ref{fig:rdp}(a)--(c), different latent paradigms favor different parts of the fidelity--perception operating space.
Distortion-oriented SQ codecs scale well at medium and high bitrates, achieving low MSE as bpp increases, but their perceptual quality drops sharply at low rates, as reflected by high LPIPS.
Perceptual SQ training improves realism at low-to-mid bitrates, yet still exhibits a distortion spike at extreme low rates ($<0.05$\,bpp), suggesting limited semantic capacity under entropy-constrained SQ latents.
By contrast, VQ-based codecs perform favorably in the ultra-low regime in terms of perceptual quality, but deliver weaker and less scalable fidelity, showing higher MSE with only modest improvement as bpp increases.
These trends indicate that SQ and VQ paradigms are not interchangeable solutions to the same latent role. Instead, they favor different reconstruction responsibilities.

Fig.~\ref{fig:rdp}(d) further illustrates this difference in latent capacity allocation.
Entropy-constrained SQ latents increasingly concentrate in high-probability regions as bitrate decreases, suppressing variation and yielding averaged reconstructions.
This concentration behavior largely persists even for perceptual SQ codecs: adding perceptual objectives mainly reshapes the decoder's synthesis preference, rather than changing the continuous, entropy-constrained latent paradigm itself.
In contrast, VQ tokenization better preserves high-level semantics at very low rates, but discretization discards fine-grained fidelity cues, limiting structural detail recovery and causing diminishing distortion-oriented gains as bitrate increases.

\subsection{From Latent Role Assignment to Decoder-Side Collaboration}

The analysis above suggests that the key bottleneck lies not only in how the fidelity--perception objective is tuned, but also in how reconstruction roles are assigned to latent representations.
SQ-based latents are better suited to scalable structural fidelity, whereas VQ-based tokens are better suited to compact semantic and perceptual cues under tight bit budgets.
This indicates that forcing a single latent paradigm to cover both roles can overload the representation, while directly merging heterogeneous representations can weaken their native specialization.

These observations naturally motivate a decoder-side collaborative decoding strategy with explicit latent roles.
Rather than treating SQ and VQ representations as features to be fused into a shared reconstruction pathway, it is more natural to keep their decoder-side specializations explicit: one branch acts as a fidelity-oriented expert, while the other acts as a perception-oriented expert.
The remaining challenge is then how to coordinate these experts without compromising their native behavior.
This motivates a decoder-side collaborative framework that preserves expert specialization while enabling selective cross-expert information transfer.

\begin{figure*}[!t]
    \includegraphics[width=0.98\linewidth]{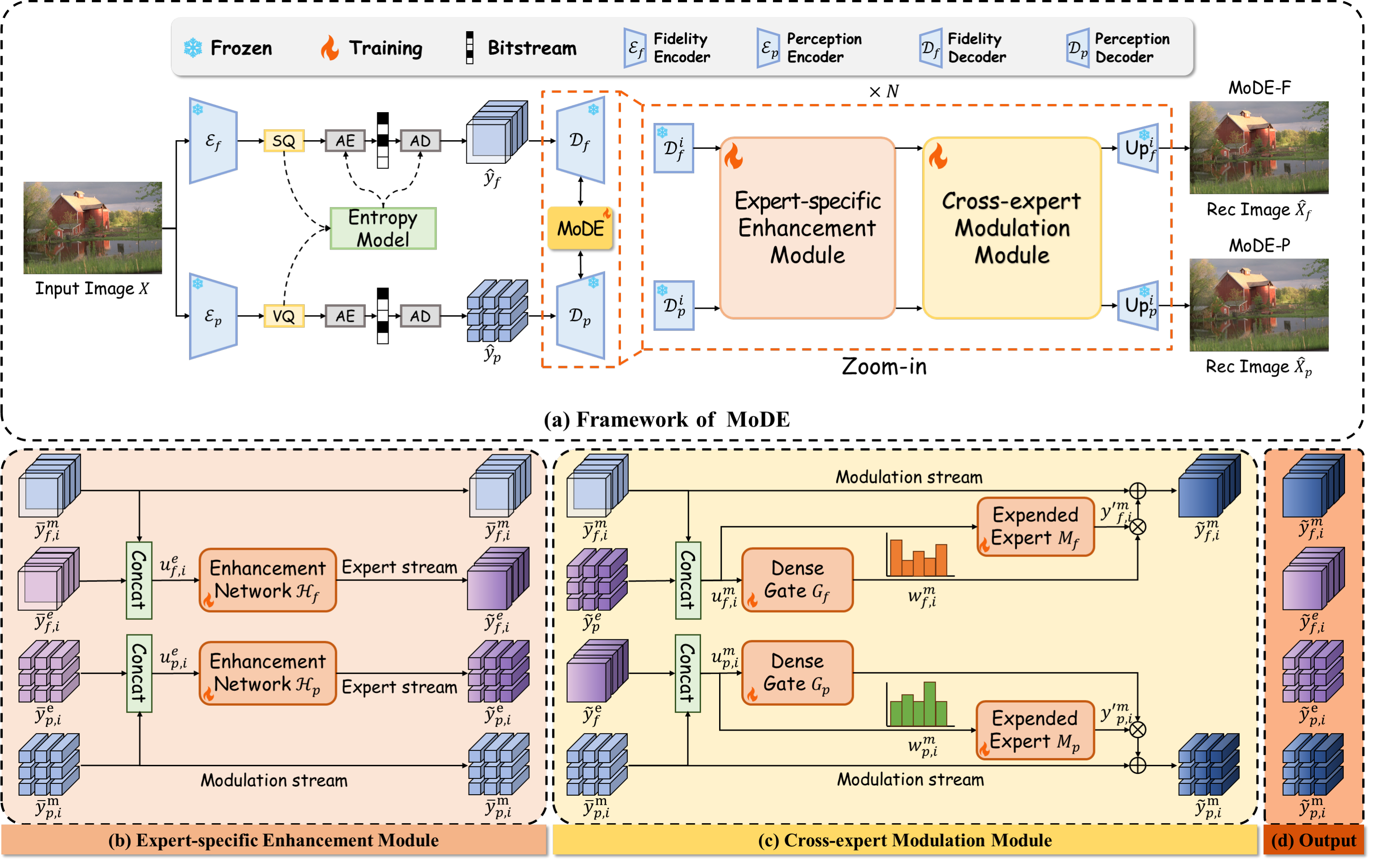}
\vspace{-3 mm}
\caption{\textbf{MoDE framework for dual-latent collaborative decoding.}
MoDE treats the SQ and VQ decoders as fidelity- and perception-oriented experts.
At each decoder level, ESE maintains expert-specific reference features, while CEM performs gated residual cross-expert modulation.
The modulation streams output two expert-anchored reconstructions, $\hat{X}^{m}_{f}$ (MoDE-F) and $\hat{X}^{m}_{p}$ (MoDE-P).}
\label{fig:framework}
\vspace{-6 mm}
\end{figure*}

\section{MoDE: Dual-Latent Collaborative Decoding via Mixture of Decoder Experts}
\label{sec:proposed-method}

\subsection{Framework Overview}
Motivated by the latent role assignment discussed above, MoDE instantiates \emph{dual-latent collaborative decoding} by treating the SQ-based decoder as a \emph{Fidelity Expert} and the VQ-based decoder as a \emph{Perception Expert}.
The Fidelity Expert provides scalable structural reconstruction, while the Perception Expert provides compact semantic and perceptual cues under tight bit budgets.
Rather than collapsing these heterogeneous representations into a single fused pathway, MoDE coordinates the two frozen experts at the decoder side, so that their complementary roles can be preserved and selectively combined during reconstruction.

As illustrated in \figref{framework}, MoDE consists of three parts:
(1) a pretrained SQ-based autoencoder optimized for distortion-oriented reconstruction,
(2) a pretrained VQ-based autoencoder equipped with a learned codebook and perceptual priors, and
(3) two decoder-side modules, \emph{Expert-Specific Enhancement (ESE)} and \emph{Cross-Expert Modulation (CEM)}.
ESE maintains branch-specific expert references to preserve specialization, while CEM selectively injects complementary cues through gated residual modulation.

\Paragraph{Encoding.}
Given an input image $X$, the two pretrained encoders produce the SQ and VQ latent representations:
\begin{equation}
\vspace{-0.5mm}
\begin{aligned}
    \hat{y}_{f} &= \mathrm{SQ}\!\big(E_{f}(X)\big), \\
    \hat{y}_{p} &= \mathrm{VQ}\!\big(E_{p}(X)\big),
\end{aligned}
\vspace{-0.5mm}
\end{equation}
where $f$ and $p$ denote the fidelity and perception branches, respectively.
$E_f(\cdot)$ and $E_p(\cdot)$ are frozen SQ- and VQ-based encoders, and $\mathrm{SQ}(\cdot)$ and $\mathrm{VQ}(\cdot)$ are the corresponding quantization operators.
Here, $\hat{y}_{p}$ denotes the quantized embedding after codebook lookup; the transmitted symbols are the corresponding discrete code indices.

\Paragraph{Decoder-side collaboration.}
MoDE is inserted into the decoding process, while both pretrained decoders remain frozen and only the decoder-side MoDE modules are learned.
For each branch $b\in\{f,p\}$, we decompose the decoder into $N$ resolution levels, where $D_b^i(\cdot)$ denotes the frozen decoding block and $\mathrm{Up}_b^i(\cdot)$ denotes the frozen upsampling block.
We use $\bar{b}$ to denote the complementary branch, \ie $\bar{f}=p$ and $\bar{p}=f$.

Each branch maintains two streams:
an \emph{expert stream} (superscript $e$), which preserves branch-specific reference features, and a \emph{modulation stream} (superscript $m$), which receives selective cross-expert cues.
Both streams are initialized from the same transmitted latent, \ie $\hat{y}^{e}_{b,0}=\hat{y}^{m}_{b,0}=\hat{y}_{b}$.
This two-stream design allows MoDE to preserve expert specialization while enabling selective decoder-side collaboration between the two branches.
\begin{equation}
\begin{aligned}
    \bar{y}^{e}_{b,i} &= D^{i}_{b}(\hat{y}^{e}_{b,i}), &
    \bar{y}^{m}_{b,i} &= D^{i}_{b}(\hat{y}^{m}_{b,i}), \\
    \tilde{y}^{e}_{b,i} &= \mathrm{ESE}_b\!\big(\bar{y}^{m}_{b,i}, \bar{y}^{e}_{b,i}\big), &
    \tilde{y}^{m}_{b,i} &= \mathrm{CEM}_b\!\big(\bar{y}^{m}_{b,i}, \tilde{y}^{e}_{\bar{b},i}\big), \\
    \hat{y}^{e}_{b,i+1} &= \mathrm{Up}_{b}^{i}\!\big(\tilde{y}^{e}_{b,i}\big), &
    \hat{y}^{m}_{b,i+1} &= \mathrm{Up}_{b}^{i}\!\big(\tilde{y}^{m}_{b,i}\big).
\end{aligned}
\label{eq:decoding_level}
\end{equation}
The expert stream provides stable branch-specific references through ESE, while the modulation stream carries the reconstruction features updated by CEM.
After $N$ levels, the modulation streams produce two expert-anchored reconstructions:
\begin{equation}
\begin{aligned}
    \hat{X}_{f}^{m} &= \mathrm{Out}_{f}\!\big(\hat{y}_{f,N}^{m}\big), \\
    \hat{X}_{p}^{m} &= \mathrm{Out}_{p}\!\big(\hat{y}_{p,N}^{m}\big),
\end{aligned}
\label{eq:decoding_output}
\end{equation}
where $\mathrm{Out}_{b}(\cdot)$ denotes the frozen output head of branch $b$.
The SQ-decoder-anchored output is denoted as \emph{MoDE-F}, which enhances perceptual realism over the fidelity expert.
The VQ-decoder-anchored output is denoted as \emph{MoDE-P}, which improves structural faithfulness over the perception expert.
The detailed designs of ESE and CEM are presented next.
%

%%%%%%%%%%%%%%%%%%%%%%%%%%%%%%%%%%
\begin{figure}[t]
    \includegraphics[width=1.0\linewidth]{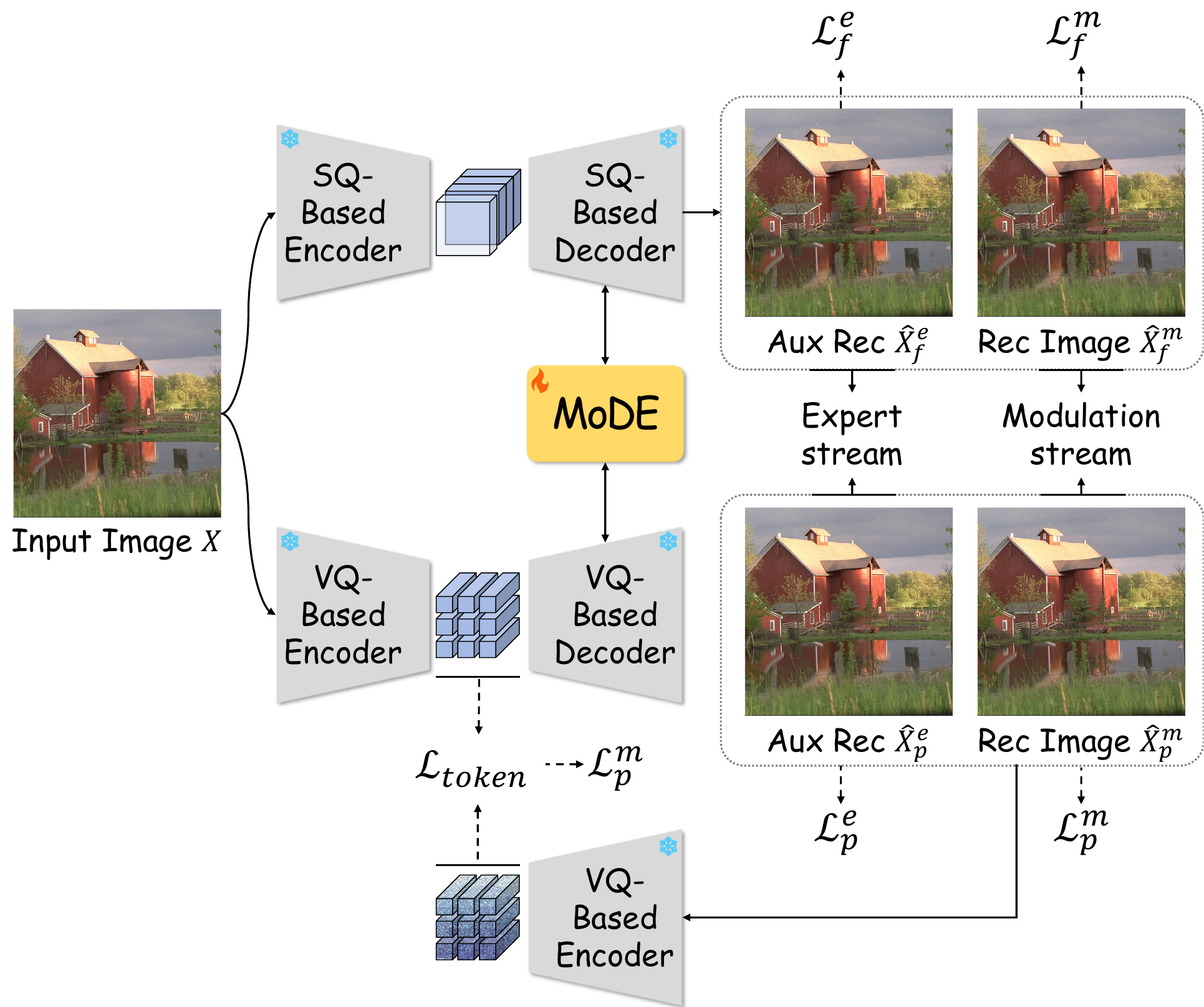}
\caption{\textbf{Training procedure of MoDE.}
Only the MoDE modules are optimized, while both pretrained SQ/VQ codecs remain frozen.
The expert stream outputs auxiliary reconstructions supervised by $(\mathcal{L}^{e}_{f}, \mathcal{L}^{e}_{p})$, and the modulation stream outputs the final reconstructions supervised by $(\mathcal{L}^{m}_{f}, \mathcal{L}^{m}_{p})$.}
    \label{fig:training}
\vspace{-6 mm}
\end{figure}
%%%%%%%%%%%%%%%%%%%%%%%%%%%%%%%%%%

\subsection{Expert-Specific Enhancement for Specialization Preservation}
\label{sec:expertenhancementmodule}

Direct cross-expert interaction can perturb the intermediate feature geometry of a frozen decoder.
For the fidelity branch, such perturbation may weaken distortion-faithful structure, whereas for the perception branch, it may reduce perceptual coherence.
Therefore, before any cross-expert transfer, MoDE first maintains a branch-specific reference stream for each expert.

To this end, we propose \emph{Expert-Specific Enhancement (ESE)}.
At each decoder level, ESE takes the modulation-stream feature and the expert-stream feature from the same branch, and outputs an enhanced expert feature that serves as a stable reference for subsequent cross-expert modulation.
Given $\bar{y}^{m}_{b,i}$ and $\bar{y}^{e}_{b,i}$, we concatenate them along the channel dimension:
\begin{equation}
    u_{b,i}^{e} = \mathrm{Concat}\!\big(\bar{y}^{m}_{b,i},\, \bar{y}^{e}_{b,i}\big),
\end{equation}
and apply an enhancement network $\mathcal{H}_b(\cdot)$:
\begin{equation}
    \tilde{y}^{e}_{b,i} = \mathcal{H}_b(u_{b,i}^{e}).
\end{equation}
ESE produces $\tilde{y}^{e}_{f,i}$ and $\tilde{y}^{e}_{p,i}$ as branch-specific expert references.
By separating the expert stream from the modulation stream, ESE preserves each frozen decoder's native specialization while providing stable reference features for CEM.
This preservation is further encouraged by the expert-preserving objective in \secref{training}.

%%%%%%%%%%%%%%%%%%%%%%%%%%%%%%%%%%
\begin{figure*}[!t]
\centering
\vspace{-5 mm}
\includegraphics[width=0.98\linewidth]{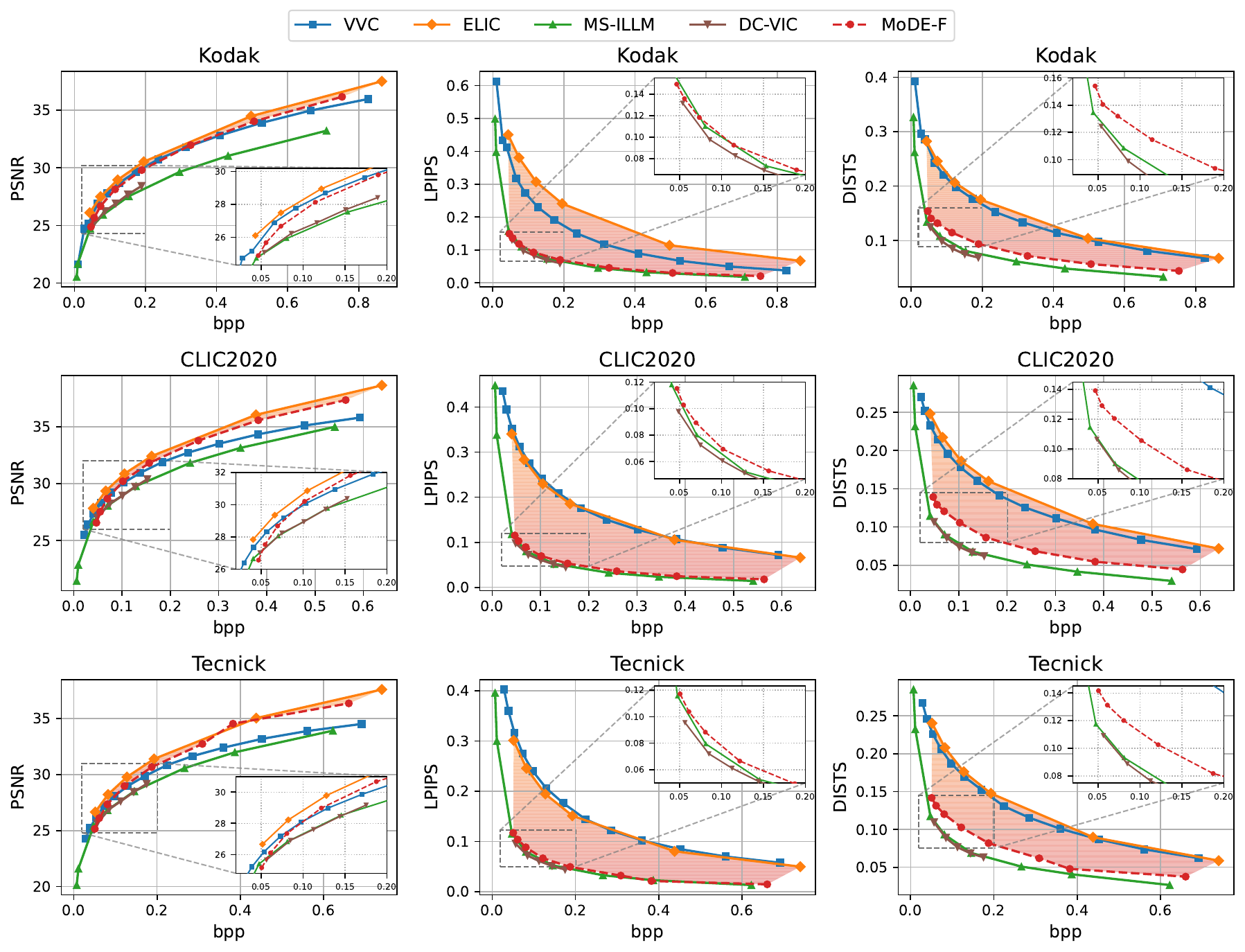}
\caption{\textbf{Rate--distortion curves on Kodak, CLIC2020, and Tecnick in terms of PSNR, LPIPS, and DISTS.} MoDE-F improves perceptual metrics with a controlled fidelity trade-off across datasets. Insets zoom into the ultra-low bitrate region.
}
\vspace{0.5mm}
\label{fig:rd_results_fidelity}
\vspace{0.5mm}
%%%%%%%%%%%%%%%%%%%%%%%%%%%%%%%%%%
\begingroup
\makeatletter
\def\@captype{table}
\makeatother
    \centering
    \caption{BD-Rate/BD-Metric comparison on the Kodak, CLIC2020, and Tecnick datasets. Anchor: ELIC~\cite{he2022elic}. %\\
    Negative BD-Rate indicates bitrate saving.
For PSNR, positive BD-metric indicates improvement;
for LPIPS/DISTS, negative BD-metric indicates improvement.
    }
    \resizebox{\linewidth}{!}{%
    \begin{tabular}{c c c c c c c}
    \toprule

    \multirow{2}{*}{\textbf{Dataset}}
    & \multirow{2}{*}{\textbf{Metric}}
    & \multicolumn{1}{c}{\textbf{Traditional Codec}}
    & \multicolumn{1}{c}{\textbf{Distortion-oriented SQ Codec}}
    & \multicolumn{1}{c}{\textbf{Perceptual SQ Codec}}
    & \multicolumn{1}{c}{\textbf{Dual-latent Fusion-based Codec}}
    & \multicolumn{1}{c}{\textbf{Dual-latent MoDE Codec}}

    \\

    \cmidrule(lr){3-3}
    \cmidrule(lr){4-4}
    \cmidrule(lr){5-5}
    \cmidrule(lr){6-6}
    \cmidrule(lr){7-7}

    &
    & \textbf{VVC}
    & \textbf{ELIC}
    & \textbf{MS-ILLM}
    & \textbf{DC-VIC}
    & \textbf{MoDE-F} \\

    \midrule

    \multirow{3}{*}{Kodak}
    & PSNR $\uparrow$ & 16.18 / -0.57 & 0.00 / 0.00 & 97.00 / -2.23 & 91.81 / -1.79 & 18.64 / -0.68 \\
    & LPIPS $\downarrow$ & -38.97 / -0.06 & 0.00 / 0.00 & -87.28 / -0.18 & -85.52 / -0.24 & -83.71 / -0.18 \\
    & DISTS $\downarrow$ & -9.31 / -0.01 & 0.00 / 0.00 & -84.99 / -0.10 & -83.33 / -0.13 & -73.07 / -0.08 \\

    \midrule

    \multirow{3}{*}{CLIC2020}
    & PSNR $\uparrow$ & 33.89 / -1.11 & 0.00 / 0.00 & 73.26 / -1.93 & 68.17 / -1.61 & 17.21 / -0.66 \\
    & LPIPS $\downarrow$ & 5.99 / 0.01 & 0.00 / 0.00 & -86.95 / -0.15 & -87.81 / -0.19 & -84.99 / -0.13 \\
    & DISTS $\downarrow$ & -14.42 / -0.01 & 0.00 / 0.00 & -85.92 / -0.10 & -85.29 / -0.12 & -73.11 / -0.07 \\

    \midrule

    \multirow{3}{*}{Tecnick}
    & PSNR $\uparrow$ & 36.81 / -1.27 & 0.00 / 0.00 & 67.17 / -1.91 & 60.55 / -1.56 & 14.58 / -0.61 \\
    & LPIPS $\downarrow$ & 10.98 / 0.01 & 0.00 / 0.00 & -83.58 / -0.12 & -82.90 / -0.16 & -79.25 / -0.11 \\
    & DISTS $\downarrow$ & -8.05 / -0.01 & 0.00 / 0.00 & -82.46 / -0.09 & -80.21 / -0.11 & -67.15 / -0.07 \\

    \bottomrule
    \end{tabular}%
    }
    \label{tab:bd_rate_elic_anchor}
\endgroup
\vspace{-2mm}
\end{figure*}

\subsection{Cross-Expert Modulation for Selective Expert Coordination}
\label{sec:dualbranchcomplementarymodulation}

With enhanced expert features from ESE, MoDE coordinates the two frozen decoders through \emph{Cross-Expert Modulation (CEM)}.
The purpose of CEM is not to uniformly fuse the two pathways, but to inject complementary cues only where they are useful.
For MoDE-F, the fidelity branch may request semantic or perceptual cues from the perception expert.
For MoDE-P, the perception branch may request structure-preserving cues from the fidelity expert.

\Paragraph{1) Dense gate prediction.}
Different from conventional Top-$K$ MoE routing~\cite{shazeer2017outrageously}, which outputs global mixture weights, CEM predicts dense branch-wise gates.
For each branch, we concatenate the modulation-stream feature with the enhanced expert feature from the complementary branch:
\begin{equation}
u_{b,i}^{m} = \mathrm{Concat}\!\big(\bar{y}_{b,i}^{m},\, \tilde{y}_{\bar{b},i}^{e}\big).
\end{equation}
The gate map is predicted by a gating network $G_b(\cdot)$:
\begin{equation}
    w^{m}_{b,i} = G_{b}\!\big(u_{b,i}^{m}\big),
\end{equation}
where $w^{m}_{b,i}$ matches the spatial resolution of the decoder feature.
A sigmoid activation constrains the gate to $[0,1]$, allowing each branch to independently control cross-expert injection.

\Paragraph{2) Branch-specific modulation and gated residual injection.}
Since the two experts require different complementary cues, CEM uses a branch-specific transform $M_b(\cdot)$ to extract modulation signals:
\begin{equation}
    {y'}^{m}_{b,i} = M_b\!\big(u_{b,i}^{m}\big).
\end{equation}
The modulation stream is then updated by gated residual injection:
\begin{equation}
    \tilde{y}^{m}_{b,i} = \bar{y}^{m}_{b,i} + w^{m}_{b,i} \odot {y'}^{m}_{b,i}.
\label{eq:modulation}
\end{equation}
This residual form keeps the original decoder feature as the anchor and uses the gate to control the strength and location of complementary transfer.
Thus, CEM enables selective cross-expert collaboration while reducing unnecessary interference between the two frozen decoders.

\subsection{Training Strategy}
\label{sec:training}

%%%%%%%%%%%%%%%%%%%%%%%%%%%%%%%%%%
\begin{figure*}[!t]
\centering
    \includegraphics[width=0.94\linewidth]{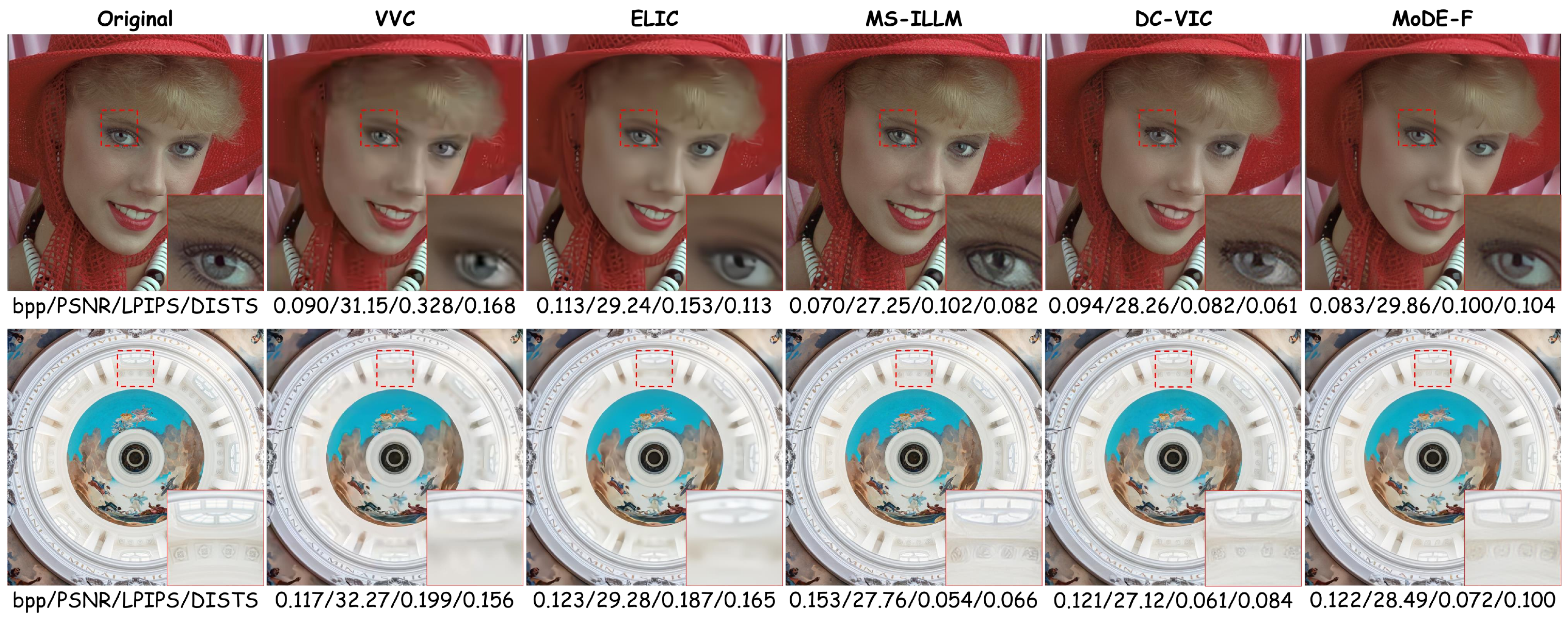}
 \vspace{-5 mm}
\caption{
\textbf{Qualitative comparisons on test images at comparable bitrates.} MoDE-F improves perceptual quality with a controlled fidelity trade-off. The bpp/PSNR/LPIPS/DISTS are reported below each image.
}
  \label{fig:fidelity_qualitative_results}
\vspace{-5 mm}
\end{figure*}
%%%%%%%%%%%%%%%%%%%%%%%%%%%%%%%%%%

As shown in \figref{training}, during training, all encoders, decoders, entropy models, and codebooks in the two expert branches remain frozen.
Only ESE and CEM are optimized.
The training objective follows the two responsibilities of MoDE:
(i) preserving expert-specific reference streams, and
(ii) learning selective complementary transfer for the modulation streams.

\Paragraph{Expert-preserving losses.}
The expert streams are supervised by the native objective of each branch:
\begin{equation}
\begin{aligned}
    \mathcal{L}^{e}_{f} &=
    \lambda^{e}_{\mathrm{MSE}}\,\operatorname{MSE}(X,\hat{X}^{e}_{f}), \\
    \mathcal{L}^{e}_{p} &=
    \lambda^{e}_{\mathrm{LPIPS}}\,\operatorname{LPIPS}(X,\hat{X}^{e}_{p}),
\end{aligned}
\label{eq:expert_loss}
\end{equation}
where $\hat{X}^{e}_{f}$ and $\hat{X}^{e}_{p}$ are auxiliary reconstructions decoded from the expert streams.
These losses encourage the fidelity expert stream to remain distortion-faithful and the perception expert stream to remain perceptually aligned.

\Paragraph{Selective-transfer losses.}
The modulation streams are supervised according to their anchored reconstruction roles.
For the perception branch, we additionally use a token-consistency loss to preserve VQ semantic consistency:
\begin{equation}
\mathcal{L}_{\mathrm{token}}
= \left\|h-\hat{h}^{m}_{p}\right\|_{1},
\quad
h=E_p(X),\quad
\hat{h}^{m}_{p}=E_p(\hat{X}^{m}_{p}).
\end{equation}
The modulation-stream objectives are
\begin{equation}
\begin{aligned}
    \mathcal{L}^{m}_{f}
    &= \lambda^{m}_{\mathrm{MSE}}\operatorname{MSE}(X,\hat{X}^{m}_{f}) + \lambda^{m}_{\mathrm{LPIPS}}\operatorname{LPIPS}(X,\hat{X}^{m}_{f}) \\
    &\quad + \lambda^{m}_{\mathrm{adv}}\mathcal{L}_{\mathrm{adv}}, \\
    \mathcal{L}^{m}_{p}
    &= \lambda^{m}_{1}\left\|X-\hat{X}^{m}_{p}\right\|_{1} + \lambda^{m}_{\mathrm{LPIPS}}\operatorname{LPIPS}(X,\hat{X}^{m}_{p}) \\
    &\quad + \lambda^{m}_{\mathrm{token}}\mathcal{L}_{\mathrm{token}} + \lambda^{m}_{\mathrm{adv}}\mathcal{L}_{\mathrm{adv}}.
\end{aligned}
\label{eq:modulation_loss}
\end{equation}
The full objective is
\begin{equation}
    \mathcal{L}_{\mathrm{total}}
    = \mathcal{L}^{e}_{f} + \mathcal{L}^{e}_{p}
    + \mathcal{L}^{m}_{f} + \mathcal{L}^{m}_{p}.
\label{eq:total_loss}
\end{equation}
The detailed training schedule and hyperparameters are provided in the supplementary material.

\section{Experiments}
\label{sec:experiments}

\subsection{Implementation Details}
\label{sec:impl_details}

We instantiate MoDE with an SQ-based distortion-oriented codec and a VQ-based perception-oriented codec, serving as the \emph{Fidelity Expert} and \emph{Perception Expert}, respectively.
For the fidelity branch, we employ ELIC~\cite{he2022elic}.
Due to the absence of an official implementation, we re-implement ELIC using the training protocol of HiFiC~\cite{mentzer2020high} and obtain comparable performance; details are provided in the \emph{Supplementary Material (SM)}.
For the perception branch, we use the Fine-tuned VQGAN from~\cite{mao2024extreme}.

In all configurations, both the SQ bitstream and the VQ code indices are transmitted.
All reported bitrates are measured in bits-per-pixel (bpp) and include the full overhead of the unified dual-stream bitstream.
To cover a broad bitrate range, we fix the VQ codebook size to $N{=}1024$ and adjust the SQ bitstream through the RD trade-off parameter.
At inference time, MoDE-F and MoDE-P use the same pretrained experts and unified dual-stream bitstream, but use their corresponding trained MoDE modules to produce fidelity-anchored and perception-anchored reconstructions.
All models are trained on a subset of OpenImages~\cite{kuznetsova2020open} using $256{\times}256$ random resized crops with batch size 6.
We further provide an additional study with a smaller VQ codebook size of $N=8$ in the \emph{SM}, showing that reducing the VQ index overhead allows MoDE to reach a lower bitrate range.

\begin{figure*}[!t]
\centering
\vspace{-6 mm}
\includegraphics[width=0.98\linewidth]{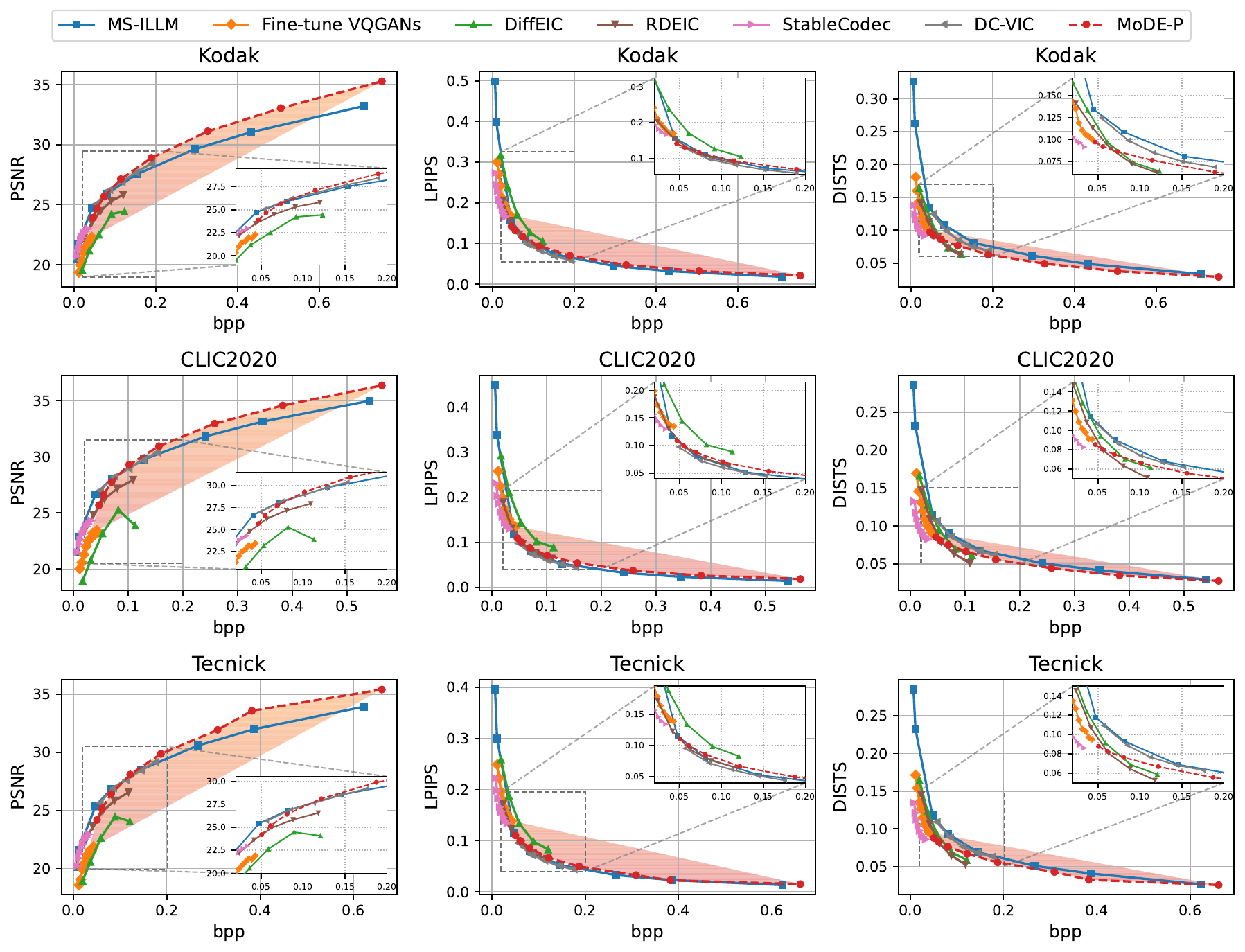}
\caption{
\textbf{Rate--distortion curves on Kodak, CLIC2020, and Tecnick in terms of PSNR, LPIPS, and DISTS.} MoDE-P extends perception-anchored reconstruction toward higher-fidelity outputs and exposes the corresponding perceptual trade-offs. Insets zoom into the ultra-low bitrate region.
}
 \vspace{-5 mm}
\label{fig:rd_results_perception}
\end{figure*}
%%%%%%%%%%%%%%%%%%%%%%%%%%%%%%%%%%

\begin{table*}[!t]
    \centering
    \caption{
    BD-rate/BD-metric comparison on Kodak, CLIC2020, and Tecnick under $<0.2$ bpp, anchored by MS-ILLM~\cite{muckley2023improving}. Negative BD-Rate indicates bitrate saving. For PSNR, positive BD-metric indicates improvement; for LPIPS/DISTS, negative BD-metric indicates improvement. We additionally report MoDE-P over the full bitrate range.}
    \resizebox{\linewidth}{!}{%
    \begin{tabular}{c c c c c c c c c c}
    \toprule

    \multirow{2}{*}{\textbf{Dataset}}
    & \multirow{2}{*}{\textbf{Metric}}
    & \multicolumn{1}{c}{\textbf{Perceptual SQ Codec}}
    & \multicolumn{1}{c}{\textbf{Perceptual VQ Codec}}
    & \multicolumn{3}{c}{\textbf{Diffusion-based Codec}}
    & \multicolumn{1}{c}{\textbf{Dual-latent Fusion-based Codec}}
    & \multicolumn{2}{c}{\textbf{Dual-latent MoDE Codec}} \\

    \cmidrule(lr){3-3}
    \cmidrule(lr){4-4}
    \cmidrule(lr){5-7}
    \cmidrule(lr){8-8}
    \cmidrule(lr){9-10}

    &
    & \textbf{MS-ILLM}
    & \textbf{Fine-tuned VQGAN}
    & \textbf{DiffEIC}
    & \textbf{RDEIC}
    & \textbf{StableCodec}
    & \textbf{DC-VIC}
    & \textbf{MoDE-P}
    & \textbf{MoDE-P (full bitrate range)}\\

    \midrule

    \multirow{3}{*}{Kodak}
    & PSNR $\uparrow$ & 0.00 / 0.00 & 228.14 / -2.65 & 302.75 / -2.95 & 76.23 / -1.12 & 46.95 / -0.88 & -5.15 / 0.13 & 9.23 / -0.11 & -12.85 / 0.63 \\
    & LPIPS $\downarrow$ & 0.00 / 0.00 & -5.96 / -0.02 & 64.78 / 0.05 & -4.88 / -0.01 & -38.21 / -0.09 & -12.53 / -0.01 & -1.88 / 0.00 &  6.10 / 0.00 \\
    & DISTS $\downarrow$ & 0.00 / 0.00 & -50.18 / -0.04 & -29.40 / -0.02 & -41.04 / -0.03 & -78.54 / -0.10 & -9.78 / -0.01 & -47.95 / -0.03 & -33.59 / -0.01 \\

    \midrule

    \multirow{3}{*}{CLIC2020}
    & PSNR $\uparrow$ & 0.00 / 0.00 & 243.26 / -3.44 & 502.99 / -4.73 & 74.28 / -1.43 & 59.29 / -1.33 & 1.32 / -0.03 &  19.61 / -0.39 & -5.36 / 0.41 \\
    & LPIPS $\downarrow$ & 0.00 / 0.00 & 10.53 / -0.00 & 85.73 / 0.05 & 9.07 / 0.00 & -29.16 / -0.07 & -9.42 / -0.00 & 9.54 / 0.01 &  14.05 / 0.01 \\
    & DISTS $\downarrow$ & 0.00 / 0.00 & -35.93 / -0.03 & -9.17 / -0.01 & -24.28 / -0.02 & -69.80 / -0.08 & -2.94 / -0.00 & -34.32 / -0.02 & -24.13 / -0.01 \\

    \midrule

    \multirow{3}{*}{Tecnick}
    & PSNR $\uparrow$ & 0.00 / 0.00 & 211.90 / -3.19 & 286.94 / -3.69 & 62.67 / -1.33 & 38.52 / -0.98 &  3.20 / -0.08 &  23.03 / -0.51 & -2.96 / 0.00 \\
    & LPIPS $\downarrow$ & 0.00 / 0.00 & 9.34 / -0.01 & 77.59 / 0.04 & 6.93 / -0.00 & -27.43 / -0.07 & -12.36 / 0.00 & 6.99 / 0.01 & 8.40 / 0.00 \\
    & DISTS $\downarrow$ & 0.00 / 0.00 & -43.34 / -0.04 & -22.93 / -0.01 & -35.08 / -0.02 & -72.81 / -0.09 & -3.93 / -0.00 & -35.24 / -0.02 & -23.52 / -0.01 \\

    \bottomrule
    \end{tabular}%
    }
    \label{tab:bd_rate_ms_illm_anchor}
\end{table*}

\subsection{Evaluation Protocol and Baselines}
\label{sec:eval_protocol}

\Paragraph{Datasets and metrics.}
We evaluate on Kodak~\cite{franzen1999kodak}, the CLIC2020 test set~\cite{CLIC2020}, and Tecnick~\cite{asuni2014testimages}.
For CLIC2020 and Tecnick, we evaluate at a fixed resolution of $768{\times}768$; the preprocessing details are provided in the SM.
We report fidelity using PSNR and perceptual quality using LPIPS~\cite{zhang2018unreasonable} and DISTS~\cite{ding2020image}.
We compute BD-Rate and BD-Metric using the standard Bj{\o}ntegaard method.

\Paragraph{Baselines.}
We compare MoDE against representative codecs covering classical, distortion-oriented, perception-oriented, generative, and dual-latent paradigms:
(a) VVC reference software (VTM-11.0)~\cite{9503377},
(b) the distortion-oriented SQ codec ELIC~\cite{he2022elic},
(c) the perception-oriented SQ codec MS-ILLM~\cite{muckley2023improving},
(d) the VQ-based perception-oriented codec Fine-tuned VQGAN~\cite{mao2024extreme},
(e) diffusion-based generative codecs DiffEIC~\cite{li2024towards}, RDEIC~\cite{li2025rdeic}, and latest StableCodec~\cite{zhang2025stablecodec}
and (f) DC-VIC~\cite{iwai2024dual}, a representative dual-latent baseline.

\Paragraph{Mode-aligned anchored evaluation.}
Since MoDE supports two decoder outputs under the same dual-stream transmission setup, we evaluate it from two anchored perspectives:
\textbf{MoDE-F} is fidelity-anchored and evaluates how perception-oriented cues improve the SQ fidelity expert, while \textbf{MoDE-P} is perception-anchored and evaluates how fidelity-oriented cues restore structure for the VQ perception expert.
Accordingly, we use mode-aligned comparisons:
\begin{compactitem}
\item \textbf{MoDE-F (ELIC-anchored).} We compare against ELIC as the anchor expert, and include VVC, MS-ILLM, and DC-VIC as cross-paradigm baselines.
\item \textbf{MoDE-P (perception-anchored).} Since Fine-tuned VQGAN is available only at sparse ultra-low bitrates, direct BD-rate measurements anchored to it are not reliable. We therefore use curve-level comparisons against Fine-tuned VQGAN and report MS-ILLM-anchored BD-rate results for consistent numerical comparison across methods.
\end{compactitem}

%%%%%%%%%%%%%%%%%%%%%%%%%%%%%%%%%%
\begin{figure*}[!t]
\centering
    \includegraphics[width=0.98\linewidth]{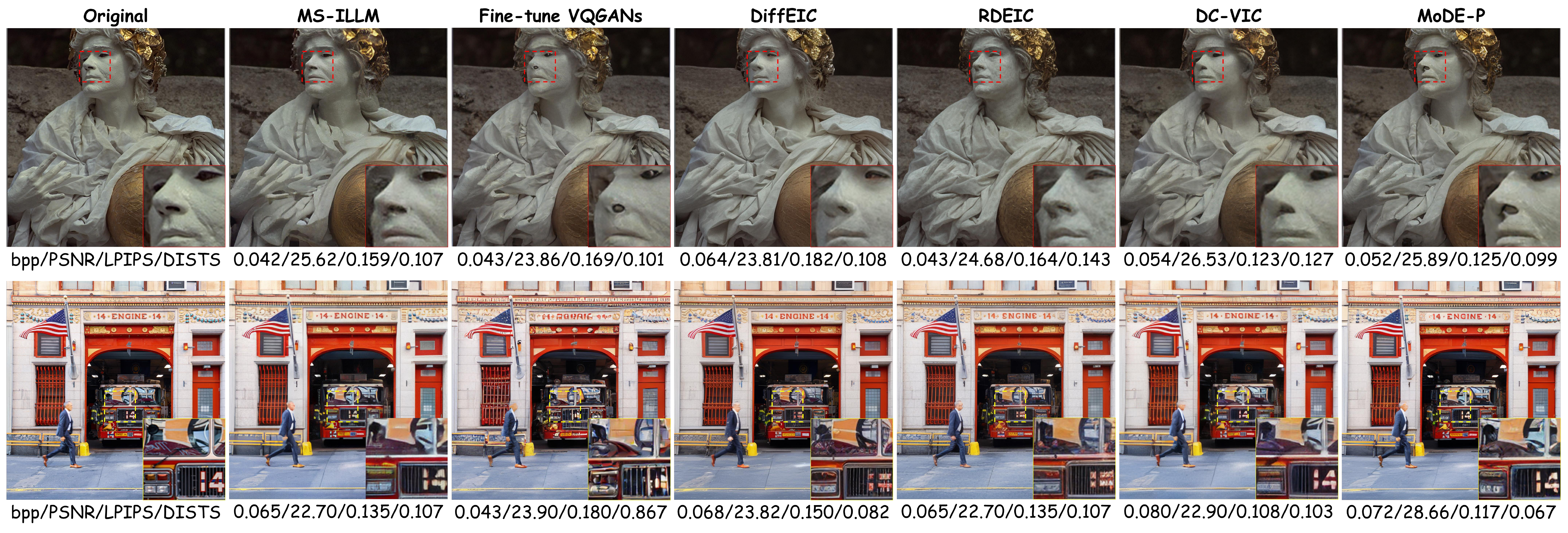}
\vspace{-4 mm}
\caption{
\textbf{Qualitative comparisons on test images at comparable bitrates.}
MoDE-P better preserves fine textures while improving reconstruction fidelity. The bpp/PSNR/LPIPS/DISTS are reported below each image.}
\label{fig:perception_qualitative_results}
\vspace{-4 mm}
\end{figure*}
%%%%%%%%%%%%%%%%%%%%%%%%%%%%%%%%%%

\subsection{Main Results: Wide-Range Fidelity--Perception Balance}
\label{sec:main_results}
We report the main results from the two anchored perspectives introduced above.
\emph{MoDE-F} evaluates perceptual enhancement over the fidelity anchor with controlled fidelity loss, while \emph{MoDE-P} evaluates structure recovery over the perception anchor under explicit perceptual trade-offs.
Together, these results assess whether decoder-side collaboration across complementary latent paradigms yields a more favorable fidelity--perception balance over a wide bitrate range.

\subsubsection{MoDE-F: Perceptual Enhancement with Competitive Fidelity}
We first evaluate \textbf{MoDE-F} using ELIC as the anchor.
As shown in \figref{rd_results_fidelity} and \tabref{bd_rate_elic_anchor}, MoDE-F improves perceptual metrics over the ELIC anchor, with large BD-rate savings in LPIPS and DISTS across Kodak, CLIC2020, and Tecnick.
These perceptual gains come with a moderate PSNR trade-off, \eg a PSNR BD-rate penalty of 18.64\% on Kodak, which is much smaller than the penalties of MS-ILLM and DC-VIC under the same ELIC anchor.
This is consistent with the design goal of MoDE-F: it injects perception-oriented cues while keeping the SQ fidelity expert as the structural anchor.

Compared with VVC, MoDE-F is competitive in PSNR at low bitrates and improves perceptual metrics more clearly.
The qualitative examples in \figref{fidelity_qualitative_results} further illustrate this behavior: traditional or distortion-oriented reconstructions tend to blur fine textures, while MoDE-F produces sharper and more visually coherent structures at comparable bitrates.
Compared with MS-ILLM and DC-VIC, MoDE-F avoids the much larger distortion penalty observed in the BD-rate table, suggesting a more controlled fidelity--perception trade-off than direct perceptual optimization or dual-latent fusion baselines.

\subsubsection{MoDE-P: Structure Recovery with Perceptual Trade-offs}
We next evaluate \textbf{MoDE-P} from the perception-anchored perspective.
Since Fine-tuned VQGAN is available only at sparse ultra-low bitrate operating points, we use it mainly for curve-level and qualitative comparison, and adopt MS-ILLM as the anchor for stable BD-rate evaluation in \tabref{bd_rate_ms_illm_anchor}.
As shown in \figref{rd_results_perception}, MoDE-P extends perception-anchored reconstruction toward higher-fidelity operating points while preserving the semantic and perceptual strengths of the VQ branch.

Relative to MS-ILLM under $<0.2$ bpp, MoDE-P consistently improves DISTS, while its PSNR and LPIPS results reflect the expected trade-off of steering the perception anchor toward more faithful structures.
When evaluated over the full bitrate range, MoDE-P achieves PSNR BD-rate savings on all three datasets and retains DISTS gains, indicating that the fidelity benefit becomes clearer beyond the ultra-low-bitrate window.
StableCodec provides favorable LPIPS/DISTS performance among diffusion-based baselines under the same MS-ILLM anchor, but still incurs positive PSNR BD-rate; by contrast, MoDE-P emphasizes structure recovery from the perception anchor rather than purely perceptual metric optimization.
Compared with diffusion-based codecs, MoDE-P better preserves input-specific structures in the qualitative comparisons.
For example, \figref{perception_qualitative_results} shows that MoDE-P preserves fine structural details that are weakened or hallucinated by some generative baselines.
These results support the intended role of MoDE-P: fidelity cues from the SQ expert help the VQ perception expert recover structure while keeping the fidelity--perception trade-off explicit.

\subsection{Ablation Study}
\label{sec:ablation_study}

\begin{table}[!t]
    \centering
    \caption{BD-rate/BD-metric of variants on Kodak, anchored by MoDE-F/MoDE-P.}
    \resizebox{0.95\linewidth}{!}{%
    \begin{tabular}{l c c c}
    \toprule
    \multicolumn{1}{c}{} & \multicolumn{3}{c}{\textbf{Kodak}} \\
    \cmidrule(lr){2-4}
    \multirow{-2}{*}{\textbf{Model Variant}} &
        \textbf{PSNR}$\uparrow$ & \textbf{LPIPS}$\downarrow$ & \textbf{DISTS}$\downarrow$ \\
    \midrule
    MoDE-F & 0.00 / 0.000 & 0.00 / 0.000 & 0.00 / 0.000 \\

    w/o ESE
        & 1.43 / -0.045
        & 8.82 / 0.004
        & 11.66 / 0.004 \\
    w/o CEM
        & 2.86 / -0.101
        & 8.84 / 0.004
        & 10.98 / 0.004 \\

    \midrule
    MoDE-P & 0.000 / 0.000 & 0.000 / 0.000 & 0.000 / 0.000 \\

    w/o ESE
        & 5.08 / -0.271
        & 3.09 / 0.002
        & 61.31 / 0.015 \\
    w/o CEM
        & 52.39 / -3.906
        & 282.42 / 0.049
        & 246.66 / 0.040 \\

    \bottomrule
    \end{tabular}}

    \label{tab:bd_metric_ablation}
\end{table}

We ablate the two key components of MoDE on Kodak: \textbf{ESE} and \textbf{CEM}.
All variants are trained and evaluated under the same unified dual-stream bitstream protocol as the full model.
The ablation is supporting evidence for the mechanism design rather than a separate claim about primary comparative performance.

\Paragraph{Effect of ESE.}
Removing ESE degrades the objective metrics for both anchored modes, with a larger impact on MoDE-P.
As shown in \tabref{bd_metric_ablation}, \texttt{w/o ESE} increases the DISTS BD-rate by 61.31\% for MoDE-P, compared with 11.66\% for MoDE-F, and also causes a larger PSNR penalty for MoDE-P.
This supports the role of ESE in maintaining stable expert references before cross-expert modulation, especially for the perception-anchored path.

\Paragraph{Effect of CEM.}
To examine the necessity of gated selective transfer, we replace CEM with an SFT-style modulation~\cite{lu2024hybridflow,iwai2024dual} that directly injects cross-branch features into intermediate decoding activations without content-adaptive gating.
As shown in \tabref{bd_metric_ablation}, this replacement causes mild but consistent degradation for MoDE-F, while the degradation is severe for MoDE-P.
The result indicates that direct feature injection is insufficient for stable perception-anchored reconstruction, and that gated residual modulation is important for controlled cross-expert transfer.

\section{Conclusion}
\label{sec:conclusions}

We presented \textbf{MoDE}, a dual-latent collaborative decoding framework for learned image compression, instantiated through a \emph{Mixture of Decoder Experts} design.
By treating an SQ-based decoder as a \emph{fidelity expert} and a VQ-based decoder as a \emph{perception expert}, and by optimizing only the decoder-side \emph{Expert-Specific Enhancement} (ESE) and \emph{Cross-Expert Modulation} (CEM) modules, MoDE preserves expert specialization while enabling selective exchange of complementary cues under a shared dual-stream bitstream.
Extensive experiments across multiple datasets and bitrate regimes show that MoDE achieves a more favorable fidelity--perception balance than representative distortion-oriented, perception-oriented, generative, and dual-latent baselines over a wide bitrate range.
Together with the ablation results, these findings highlight decoder-side collaboration across complementary latent paradigms as an effective design for relieving single-latent overload in learned image compression.
\renewcommand{\IEEEbibitemsep}{0pt plus 0.5pt}
\setlength{\itemsep}{-0.2em}
\setlength{\parsep}{0pt}
\setlength{\parskip}{0pt}

\bibliographystyle{IEEEbib}
\bibliography{refs}

 \end{document}